\documentclass[letterpaper, 10 pt, conference]{ieeeconf}

\IEEEoverridecommandlockouts
\usepackage[pdftex]{graphicx}
\usepackage{caption}

\makeatletter
\let\MYcaption\@makecaption
\makeatother
\usepackage[font=footnotesize]{subcaption}
\makeatletter
\let\@makecaption\MYcaption
\makeatother
\DeclareCaptionLabelSeparator{periodspace}{.\quad}
\usepackage{algorithmic}
\usepackage{algorithm}
\usepackage{amssymb,amsmath}
\usepackage{roboticsMathNotations}
\usepackage[table,usenames,svgnames]{xcolor}
\definecolor{lightgray}{gray}{0.9}
\usepackage{multicol}
\usepackage{multirow}
\usepackage{booktabs}
\usepackage[utf8x]{inputenc}
\usepackage[english]{babel}
\usepackage[tight-spacing=true,range-units=single,separate-uncertainty=true]{siunitx}
\usepackage{pgf}
\usepackage{pgfplots}
\usepackage{pgfplotstable}
\pgfplotsset{compat=newest}
\newcommand{\subparagraph}{} 
\usepackage[explicit]{titlesec} 
\usepackage{rotating}
\usepackage{longtable}
\usepackage{colortbl}

\usepackage{tikz}

\newcommand{\frameRobot}{R}
\newcommand{\frameOdom}{O}
\newcommand{\frameCamera}{C}
\newcommand{\frameMarker}{M}
\newcommand{\framePanel}{P}
\newcommand{\robotPoseSimulation}{\ensuremath{\trfFr{\frameRobot}{\frameOdom}{_{S}}}}
\newcommand{\robotPoseMarker}{\ensuremath{\trfFr{\frameRobot}{\frameOdom}{_{M}}}}
\newcommand{\robotPoseFilter}{\ensuremath{\trfFr{\frameRobot}{\frameOdom}{_{F}}}}
\newcommand{\panelPoseSimulation}{\ensuremath{\trfFr{\framePanel}{\frameOdom}{_{S}}}}
\newcommand{\panelPoseMarker}{\ensuremath{\trfFr{\framePanel}{\frameOdom}{_{M}}}}

\newcommand{\robotPoseError}[2]{\ensuremath{d\left(\trfFr{\frameRobot}{\frameOdom}{_{#1}},\trfFr{\frameRobot}{\frameOdom}{_{#2}}\right)}}
\newcommand{\robotPositionError}[2]{\ensuremath{d\left(\posFrMean{\frameRobot}{\frameOdom}{_{#1}},\posFrMean{\frameRobot}{\frameOdom}{_{#2}}\right)}}
\newcommand{\robotOrientationError}[2]{\ensuremath{d\left(\quatFrMean{\frameRobot}{\frameOdom}{_{#1}},\quatFrMean{\frameRobot}{\frameOdom}{_{#2}}\right)}}
\newcommand{\panelPoseError}[2]{\ensuremath{d\left(\trfFr{\framePanel}{\frameOdom}{_{#1}},\trfFr{\framePanel}{\frameOdom}{_{#2}}\right)}}
\newcommand{\panelPositionError}[2]{\ensuremath{d\left(\posFrMean{\framePanel}{\frameOdom}{_{#1}},\posFrMean{\framePanel}{\frameOdom}{_{#2}}\right)}}
\newcommand{\panelOrientationError}[2]{\ensuremath{d\left(\quatFrMean{\framePanel}{\frameOdom}{_{#1}},\quatFrMean{\framePanel}{\frameOdom}{_{#2}}\right)}}

\usepackage{url}

\makeatletter
\newcommand{\todolist}[1]{\begin{itemize}\item \textcolor{red}{#1}\checknextarg}
\newcommand{\checknextarg}{\@ifnextchar\bgroup{\gobblenextarg}{\end{itemize}}}
\newcommand{\gobblenextarg}[1]{ \item \textcolor{red}{#1}\@ifnextchar\bgroup{\gobblenextarg}{\end{itemize}}}

\newcommand\inputpgf[2]{{
\let\pgfimageWithoutPath\pgfimage
\renewcommand{\pgfimage}[2][]{\pgfimageWithoutPath[##1]{#1/##2}}
\input{#1/#2}
}}

\usepackage{tabularx}

\usepackage{adjustbox}

\title{\bf Robust Continuous System Integration\\for Critical Deep-Sea Robot Operations\\Using Knowledge-Enabled Simulation in the Loop}
\author{Christian A. Mueller\textsuperscript{*}, Tobias Doernbach\textsuperscript{*}, Arturo Gomez Chavez\textsuperscript{*}, Daniel K\"ohntopp\textsuperscript{*}, Andreas Birk 
\thanks{\textsuperscript{*}These authors contributed equally to this work and share first authorship. The authors are with the Robotics Group, Computer Science \& Electrical Engineering, Jacobs University Bremen, Germany, \texttt{\{chr.mueller,\allowbreak t.fromm,\allowbreak a.gomezchavez,\allowbreak d.koehntopp,\allowbreak a.birk\}@jacobs-university.de}.}
\thanks{The research leading to the presented results has received funding from the European Union's Horizon 2020 Framework Programme (H2020-EU.3.2.) within the project (ref.: 635491) ``Effective dexterous ROV operations in presence of communication latencies (DexROV)''. The authors would like to thank the DexROV team within the Space and Innovations Division of COMEX S.A., Marseille, France, for their support in collecting the experimental data described in this paper.}
}
\begin{document}
\maketitle

\begin{abstract}
Deep-sea robot operations demand a high level of safety, efficiency and reliability.
As a consequence, measures within the development stage have to be implemented to extensively evaluate and benchmark system components ranging from data acquisition, perception and localization to control.
We present an approach based on high-fidelity simulation that embeds spatial and environmental conditions from recorded real-world data. 
This \emph{simulation in the loop} (SIL) methodology allows for mitigating the discrepancy between simulation and real-world conditions, e.g. regarding sensor noise.
As a result, this work provides a platform to thoroughly investigate and benchmark behaviors of system components concurrently under real and simulated conditions.
The conducted evaluation shows the benefit of the proposed work in tasks related to perception and self-localization under changing spatial and environmental conditions.
\end{abstract}

\section{Introduction}\label{sec:intro}
The rapid progress of Unmanned Underwater Vehicle (UUV)  capabilities in recent years has increased their use in inspection and mapping activities as they offer higher data transmission rates through acoustics, more accurate navigation and denser environment 3D models with energy-efficient sensors. 
UUV systems are also increasingly applied to areas which are inaccessible and hazardous to humans. For example, the UK Health \& Safety Executive's (HSE) 2015--2016  Offshore  Safety  Statistics~\cite{HSE2016} reports 53 major injuries and more than 400 dangerous occurrences, most of them performing maintenance and construction activities. 
However, the development and continuous evaluation of such underwater robotic systems typically requires the organization of a crew (intendant, operator, navigator) and an adequately equipped vessel to deploy, operate, and retrieve the robot offshore. This rapidly increases the effort and cost of each development cycle. 
Specialized testing and deployment strategies are required because effort and costs in case of failure are higher by several orders of magnitude than in ground robotics, for example, when a UUV malfunctions in deep sea and cannot be retrieved anymore.
Consequently, efficient strategies have to be incorporated to validate effectiveness, robustness and reliability of the developed capabilities.
As to alleviate these costs and efforts, we propose a methodology that uses a simulator for underwater robotic activities and integrates parts of the development stack with real-world data recorded from field trials.

In the context of the EU-funded research project \emph{DexROV} (Effective Dexterous ROV Operations in Presence of Communication Latencies \cite{Gancet2016}), in our previous work \cite{Fromm2017} we proposed a versatile integration and validation architecture that allows for pre-deployment testing using simulated and real system components besides each other in a seamless way. 
That work focused on the \emph{continuous system integration and deployment} of a fully integrated system that may contain simulated components due to their developmental stage.

\begin{figure}[t]
	\centering
	\includegraphics[width=0.8\linewidth]{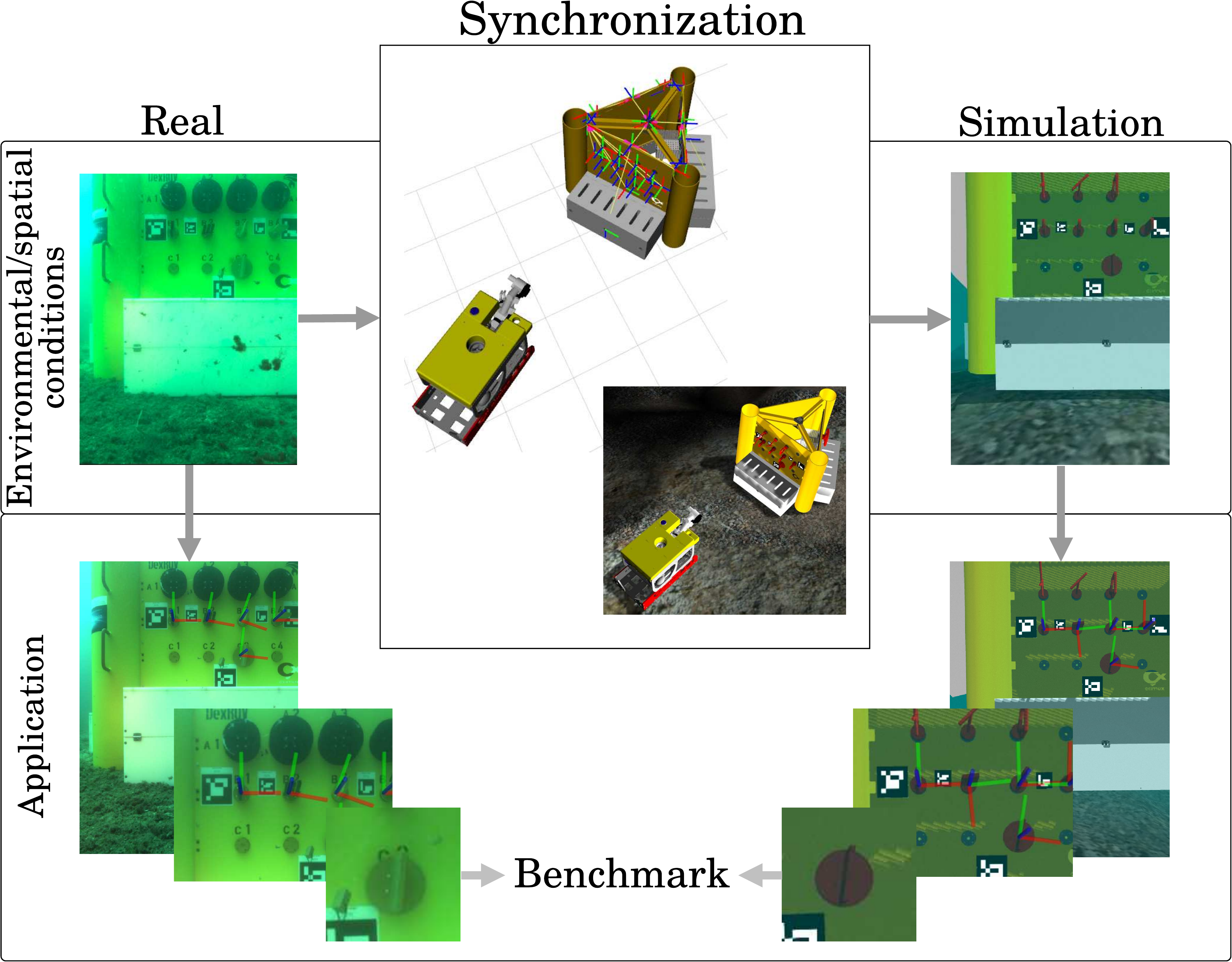} 
	\caption{Illustration of the proposed \emph{simulation in the loop} (SIL) methodology showing a perception task application (valve and lever pose estimation, see Section~\ref{sec:method:panel_comp_det}) within the DexROV project~\cite{Gancet2016}. 
	\newline \textbf{Supplementary video:} {http://robotics.jacobs-university.de/videos/sil-2018}}
	\label{fig:sim_loop}
\end{figure}

In the work presented here, however, our goal is a \emph{simulation in the loop} (SIL) architecture (see Fig.~\ref{fig:sim_loop}) which allows for extensive \emph{system benchmarking}.
Hence, our particular focus is set on closing the discrepancy between simulated and real-world data. As a result, our proposed framework
\begin{itemize}
	\item \emph{synchronizes} simulated and real-world data by incorporating environmental and spatial feedback collected from field-trials which 
	\item \emph{provides} an augmented virtual environment reflecting environmental/spatial conditions from real-world missions to test, benchmark and compare behaviors of system modules,
	\item \emph{preserves} the benefits of continuous system integration to perform such benchmarks using real or simulated components or a combination of both, and allows to
	\item \emph{perform} tests on distributed deployment, interfaces/pipelines, data regression/degradation, and fault recovery/safety as described in \cite{Fromm2017}.
\end{itemize}

\section{Related Work}\label{sec:related_work}
DexROV~\cite{Gancet2016} features a full-fledged Unmanned Underwater Vehicle (UUV) system, deployed from a vessel in the Mediterranean to perform perception as well as dexterous manipulation. In our application, an artificial testing panel has been set up to allow for interventions including, but not limited to, visual inspection, docking, and manipulation of valves, handles and other movable parts.
Such robotic underwater operations require accurate pose estimation of task-related objects like levers and valves to reduce the risk of costly failures.
Hence, robust detection of spatio-temporal reference points is of particular interest during these tasks.

Due to the noisy nature of underwater scenarios and the precision required for manipulation tasks, we exploit a-priori knowledge about the environment.
Using known landmarks allows the system to operate with low-quality data which commonly appears in deep-sea sensing, e.g.\ acoustic sensors affected by salinity and temperature, and camera images distorted by light backscatter. 
Such landmarks have been used for navigation and docking of underwater vehicles \cite{Hildebrandt2017_docking,Murarka2009_VisionbasedFS} since underwater no global positioning information is available. Additionally, artificial structure-based perception has been frequently used both in ground \cite{Salas2013_SLAM} and underwater robotics \cite{Ribas2008_UWSLAM}, \cite{Englot2010_marineinspection}. 
For these reasons, we equipped the described testing panel with visual markers~\cite{GarridoJurado2014} to exploit it as a reference structure for enhancing localization and object perception capabilities.

The major goal of the proposed SIL methodology is robust projection of real conditions to simulation, including task-related objects.
Inferences from simulated interaction of robots with the environment is a well-established approach \cite{Fromm2016a,Battaglia2013,Kunze2015}. Additionally, several frameworks exist which allow for the reproduction of experiments from a knowledge base incorporating experimental data and information deduced thereof \cite{Beetz2015,Lier2016}.
However, synchronizing the state of the simulation with real-world data for reasoning and benchmarking has still not been covered extensively in the literature.

Although numerous sophisticated simulators exist for ground and aerial robots, there is a limited number for underwater applications \cite{Prats2012_UWSim,Kermorgant2014} due to the difficulty of modeling hydrodynamic forces and environment light. 
Closest to the DexROV scenario is the UUV Simulator package \cite{Manhaes2016} which serves similar use cases like intervention tasks using a UUV manipulator. 
This package, as well as our approach, builds upon the established Gazebo \cite{Koenig2004} simulator.
Further on, to guarantee authentic projection of real conditions to simulation, we exploit visual markers attached to the testing panel as reference points, described in Section~\ref{sec:sim_real_loop}.
The work in \cite{Santos2015_UWmarkers} and our experimental evaluation in Section~\ref{exp:panel_detection} validate the use of visual markers subject to underwater image distortions.

In the remainder of this paper, we describe how to use artificial man-made structures to set up the simulation environment for perception and manipulation tasks and interlock it with continuous system integration.
In Section~\ref{sec:experiment}, we show the performance improvements achieved through experiments in object pose estimation and robot localization as real data is iteratively used to refine the simulated scenario.

\section{Simulation for\\Continuous System Integration}
\label{sec:method:cont_integ}

In a real-world robotic system, a large number of system integration hours have to be invested before being able to deploy it into a realistic scenario.
Regarding the example of DexROV, the research project features an fully-fledged UUV for open sea usage and, amongst others, equipped with a stereo perception system and a manipulator. 
Within this project, several real-world field trials have been scheduled and partly performed already.
However, the different workgroups do not have access to the hardware prior to the annual trials despite the need to thoroughly test their developed hardware and software components.

For this reason, high priority was assigned to establish a simulation base before implementing any task-specific software modules in order to allow developers to work independently and in parallel. The architecture of such a simulation framework depends on the application scenario, though the underlying principles and procedures of continuous system integration as described in detail in our previous work \cite{Fromm2017} translate to any desired area of robot operation.

\begin{figure}[t]
	\centering
	\includegraphics[width=0.9\linewidth]{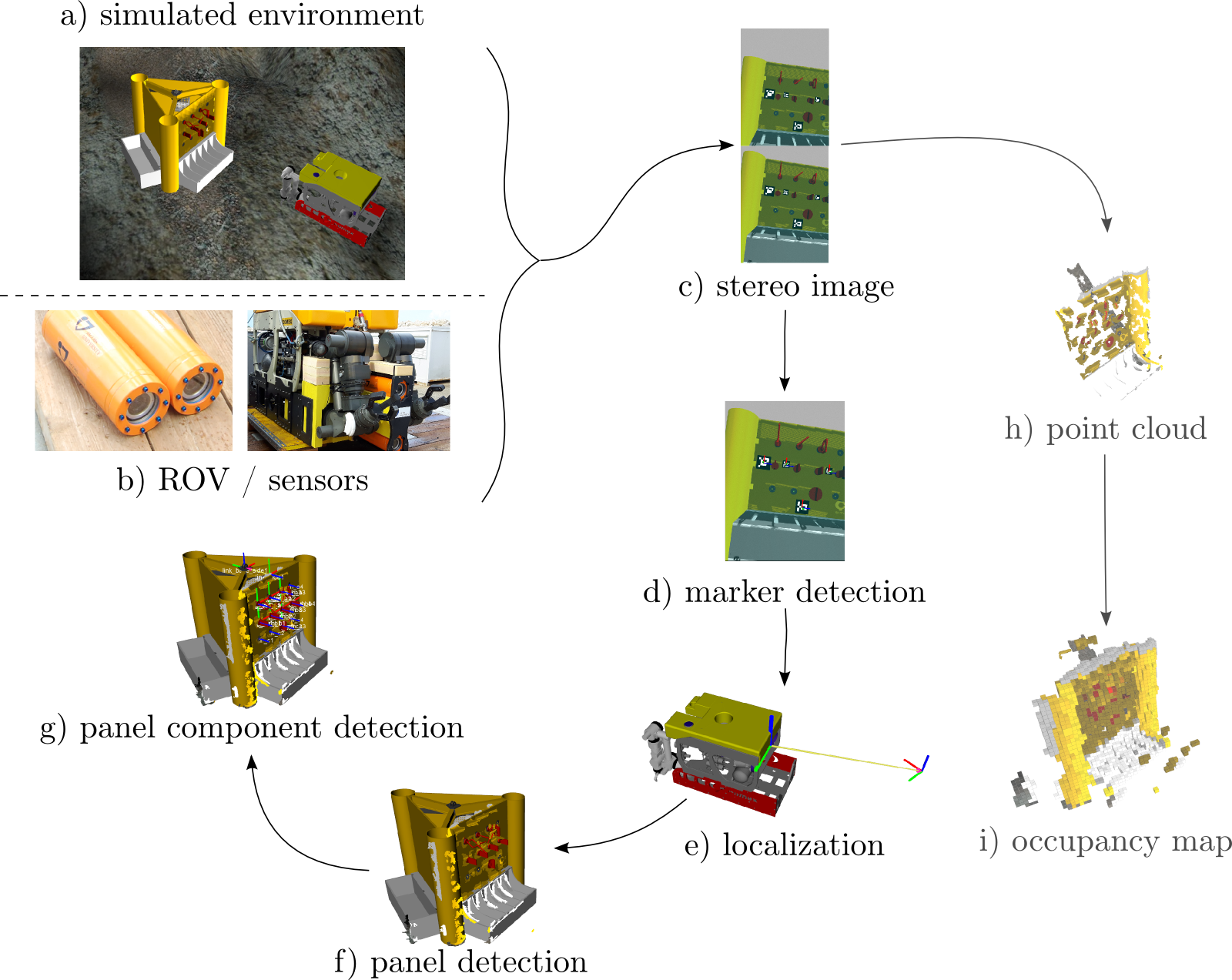} 
	\caption{System Architecture Overview -- subfigures h) and i) are not regarded in the context of this paper, see \cite{Fromm2017} instead.}
	\label{fig:overview}
\end{figure}

Fig.~\ref{fig:overview}(a) shows an example of the DexROV simulation environment including a man-made testing panel which features different valves and levers, specifically generated to test the manipulation capabilities of the robot. 
The main target of using simulation in a continuous system integration fashion is to validate a plethora of different components in simulation and to infer from their behavior onto real-world missions.
Hence, the simulation framework for the respective application scenario, i.e.\ a deep-sea setting like DexROV, needs to provide its capabilities in a component-wise structure where simulated can be effortlessly replaced with real elements.
As for inter-component communication, since such a variable system encompasses many interacting modules, all components are integrated into a middleware like ROS. This allows operability through remote and unstable network connections as described in \cite{Gancet2016} and required in harsh deployment environments.

Using the principle of continuous system integration, a complex system architecture can be established, maintained, and exploited like the application example in Fig.~\ref{fig:overview}. All the components pictured therein have been developed, enhanced, and evaluated following this schema. In Section~\ref{sec:app}, we describe in detail the business logic of some components as well as how the simulation framework has been exploited to improve their utility and usability.
However, first of all, the next section explains how real-world data is incorporated into the simulation environment for fast component tuning that, in turn, yields high-accuracy results in real-world tests.

\section{Increasing Continuous System Integration Robustness Using Simulation in the Loop}
\label{sec:sim_real_loop}
A developmental procedure that incorporates a simulation of the application scenario is a powerful tool to decrease the overall project costs by accelerating the planning and execution of field trials. It also increases the system reliability by benchmarking its behavior in extreme border conditions and without exposing valuable equipment to danger.

The concept of \emph{simulation in the loop} (SIL) goes one step further.
Instead of conducting working cycles in a sequential development-evaluation process with data generated from simulation and later from real-world field trials, SIL aims to combine both steps. 
The recorded real-world data is projected into the simulation environment using similar conditions such as the robot configuration in space, perceived sensor data and environmental constraints.

\begin{figure*}
	\centering	
    \subcaptionbox{Real-world camera image\label{fig:sim_loop_panel_det_real}}{\includegraphics[height=0.15\linewidth]{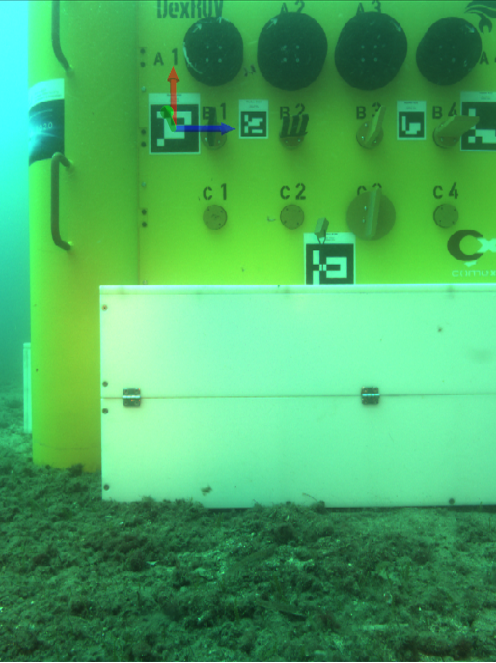}} \quad
    \subcaptionbox{Robot--panel space transformations\label{fig:rov_panel_tf}}{\includegraphics[height=0.2\linewidth]{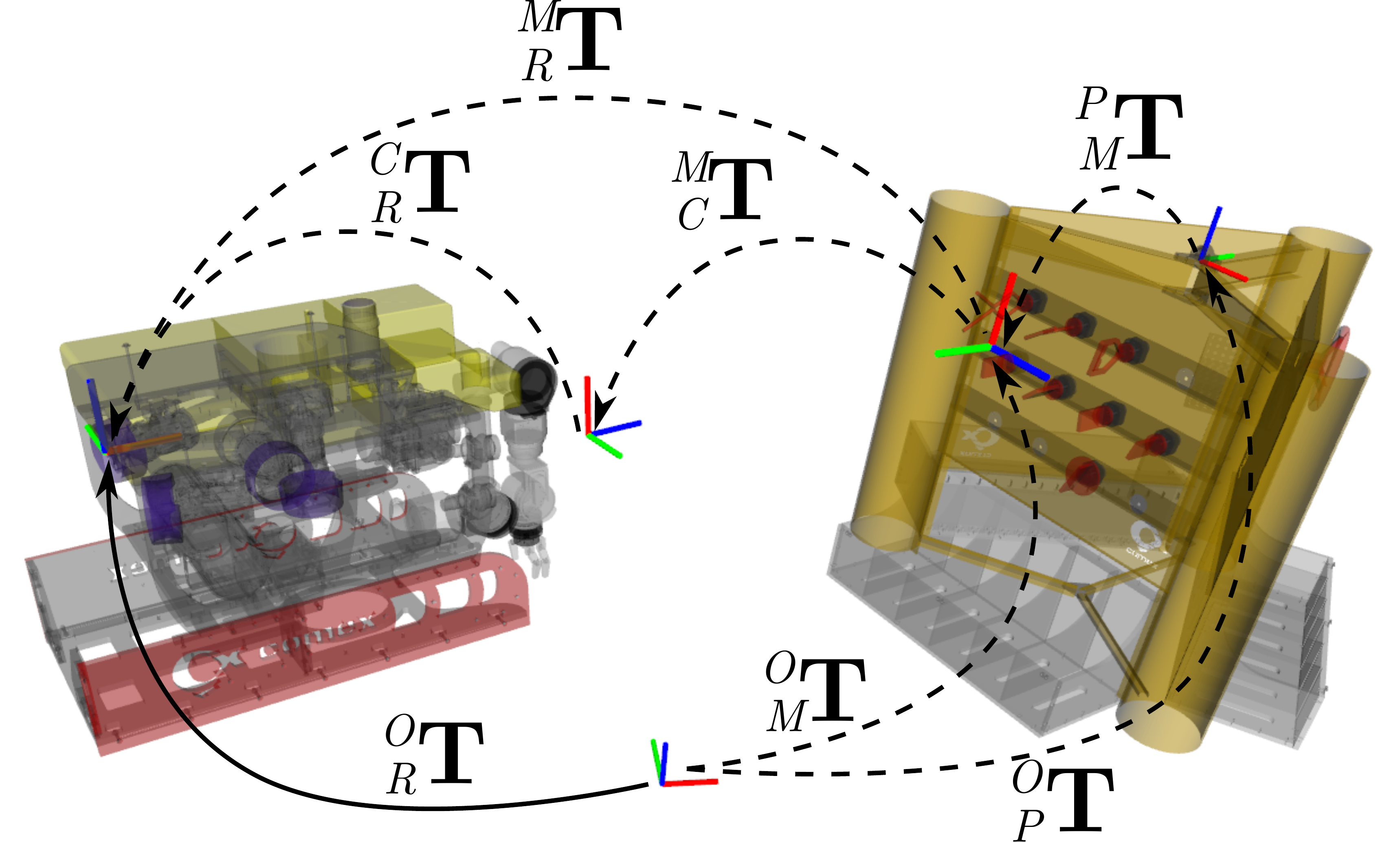}} \quad
    \subcaptionbox{Simulated camera image\label{fig:sim_loop_panel_det_sim_view}}{\includegraphics[height=0.15\linewidth]{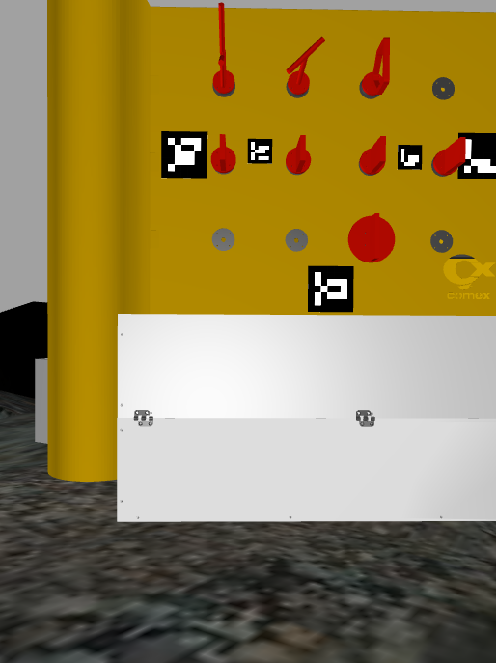}} \quad
    \subcaptionbox{Adapted simulated camera image\label{fig:sim_loop_panel_det_sim_view_adapted}}{\includegraphics[height=0.15\linewidth]{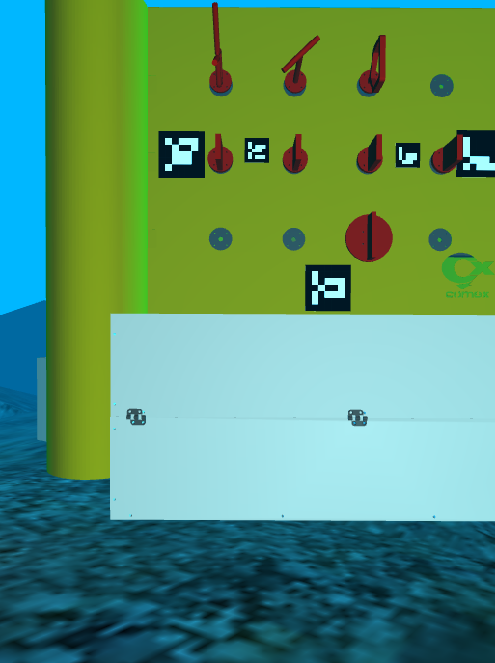}} \quad
    \subcaptionbox{Panel with projected kinematic model\label{fig:sim_loop_panel_det_km}}{\includegraphics[height=0.15\linewidth]{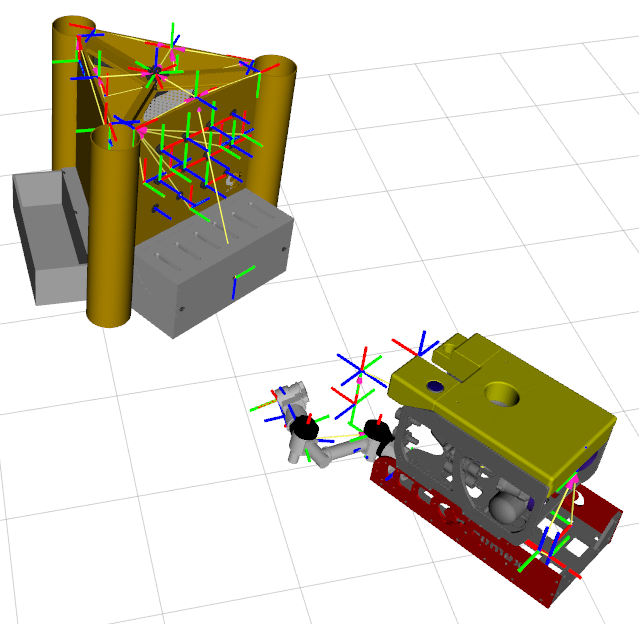}} \quad
	\caption{(\emph{Spatial condition}) Synchronization of observed robot pose in real-world data (\subref{fig:sim_loop_panel_det_real}, \subref{fig:rov_panel_tf}) with simulation environment (\subref{fig:sim_loop_panel_det_sim_view}). (\emph{Environmental condition}) Simulated images are shown in (\subref{fig:sim_loop_panel_det_sim_view}) and (\subref{fig:sim_loop_panel_det_sim_view_adapted}) accordingly.
	(\subref{fig:sim_loop_panel_det_km}) Panel component kinematic state projected in simulation. }
	\label{fig:sim_loop_panel_det}
\end{figure*}

To conflate the \emph{spatial conditions} present in real-world data with simulation, spatio-temporal reference points are used during the field trials. 
Crucial prerequisite of these reference points is their accurate detection and pose estimation in 3D space.
In the DexROV scenario, the testing panel is exploited as such a reference point since the pose estimation of the panel is a major project objective, further described in Section~\ref{sec:method:panel_det}.
To guarantee robust pose estimation, the panel is augmented with visual markers, specifically ArUco markers \cite{GarridoJurado2014} which provide high pose accuracy.
Given the augmented panel model, the observation of markers in the recorded real-world data allows to take the panel as a \emph{visual landmark} and infer the robot pose with respect to it.
This inference allows to project the relative spatial circumstances between panel and robot into simulation (Fig.~\ref{fig:rov_panel_tf}).
Likewise, states of panel components (e.g. valves, switches or wheels) from real observations can also be accordingly projected to panel components in simulation (Fig.~\ref{fig:sim_loop_panel_det_km}).

Furthermore, real observed \emph{environment conditions}, like camera image noise, haze or illumination, can be estimated 
to reflect similar conditions in simulation. For this, we rely on Gazebo's built-in sensor noise models, scene fog and lighting options as well as the UUV Simulator camera plugin \cite{Manhaes2016}.
Consequently, \emph{spatial} and \emph{environment conditions} perceived from real observations are continuously reflected in simulation in a cylic manner, as shown in Fig.~\ref{fig:sim_loop_panel_det}. This \emph{processing loop} of real observations acquisition and their projection into simulation is shown in Alg.~\ref{alg:sim-loop}.

\begin{algorithm}
	\scriptsize
	\caption{\small Simulation in the loop (SIL)}
	\label{alg:sim-loop}   
	\begin{algorithmic}
		\floatname{algorithm}{Procedure}
		\renewcommand{\algorithmicrequire}{\textbf{Input:}}
		\renewcommand{\algorithmicensure}{\textbf{Output:}}
		\REQUIRE real-world sensor data $\mathcal{R}(\mathcal{E})$, task $\mathcal{T}$
		\STATE initialize knowledge base (see Section~\ref{sec:app})
        \STATE infer environment conditions $\mathcal{E}$ from sensor data $\mathcal{R}(\mathcal{E})$
		\STATE initialize simulation environment, spawn testing panel model
		\FORALL{real-world samples $r(t) \in \mathcal{R}$ at time steps $t$}
		\STATE detect visual markers in $r(t)$ (see Fig.~\ref{fig:sim_loop_panel_det_real}) 
		\STATE estimate panel pose in odometry frame $\trfFr{\framePanel}{\frameOdom}$ from marker poses (Section~\ref{sec:method:panel_det})
		\STATE infer robot pose in odometry frame $\trfFr{\frameRobot}{\frameOdom}$ (see Fig.~\ref{fig:rov_panel_tf})
		\STATE set robot pose in simulation according to $\trfFr{\frameRobot}{\frameOdom}$ (see Fig.~\ref{fig:sim_loop_panel_det_sim_view})
		\STATE generate simulated sensor data $s(t) \in \mathcal{S}(\mathcal{E})$
		\STATE create benchmarking sensor data pair $b(t) \gets \langle r(t),s(t)\rangle$ (Figs.~\ref{fig:sim_loop_panel_det_real} and \ref{fig:sim_loop_panel_det_sim_view_adapted})
		\STATE calculate measure $m(\mathcal{T},t)$ using $b(t)$ under $\mathcal{R}(\mathcal{E})/\mathcal{S}(\mathcal{E})$ domain (Section~\ref{sec:app})
		\STATE $\mathcal{B} \gets \mathcal{B} \cup b(t)$, $M \gets M \cup m(\mathcal{T},t)$
		\ENDFOR
		\ENSURE synchronized benchmarking data sequence $\mathcal{B} \gets \mathcal{R}(\mathcal{E}) \cap \mathcal{S}(\mathcal{E})$, sequence of measures $M(\mathcal{T})$ for task $\mathcal{T}$
	\end{algorithmic}
\end{algorithm}

Through this loop process several specialized optimization and benchmarking tasks $\mathcal{T}$ can be performed iteratively based on simulated $\mathcal{S}$ and real-world data $\mathcal{R}$, such as object recognition, manipulation or 3D modeling. This yields a corresponding sequence $M(\mathcal{T})$ of individual measures $m(\mathcal{T}) \in M$ which have to be defined depending on the task $\mathcal{T}$ prior to running the simulation loop. Several examples for such measures are described and used in our experimental evaluation (Section \ref{sec:experiment}).

Additionally, environmental conditions $\mathcal{E}$ in simulation can be adapted to compare the methods performance (e.g. robot localization, panel pose estimation) under various configurations with respect to their performance under real field trial conditions. For numeric optimization, a respective task-dependent measure $m(\mathcal{T})$ can be utilized. This capability of the proposed approach is particularly valuable in continuous system development under challenging and dynamic conditions, such as in deep-sea projects like DexROV.

\section{Application-Relevant Benchmarking Tasks}
\label{sec:app}
Since deep-sea missions are cost-intensive and bear a risk to life and equipment, prior knowledge about the mission decreases risk of failures and increases safety. 
Particularly in visual inspection or manipulation tasks of man-made structures, the incorporation of prior knowledge can be exploited to increase efficiency and effectiveness of conducted missions. 
Therefore, a \emph{knowledge base} is built which contains properties of task-related objects. 
Along with offline information, like CAD models and kinematic descriptions of the robot and testing panel, the knowledge base also contains online state information gathered over the execution course of the task, e.g.\ the current robot and object poses.

Using this prior knowledge and online state information, a multitude of different validation and optimization tasks can be carried out with the presented setup.
In this section, we describe several application scenarios where the SIL methodology provides significant benefits for validation and benchmarking. All these benchmarking tasks $\mathcal{T}_P$, $\mathcal{T}_H$ and $\mathcal{T}_L$ are described in detail while their respective results can be found in our experimental evaluation.

\subsection{Panel Pose Estimation ($\mathcal{T}_P$)}
\label{sec:method:panel_det}
Our panel pose estimation approach described below is the basis for projecting the panel model and its kinematic properties into the simulation as illustrated in Fig.~\ref{fig:sim_loop_panel_det_km}.
The estimation of accurate panel poses is crucial for reliable manipulation of valves and handles.
Our approach incorporates offline knowledge such as the panel CAD model and visual markers placed at predefined locations.	
Based on this augmentation of the panel with markers, the panel pose in odometry frame $\trfFr{\framePanel}{\frameOdom}$ can be reliably estimated using the detected markers poses w.r.t.\ the camera frame $\trfFr{\frameMarker}{\frameCamera}$, the camera pose on the robot frame $\trfFr{\frameCamera}{\frameRobot}$ and the current robot pose in odometry frame \trfFr{\frameRobot}{\frameOdom}, see Fig.~\ref{fig:rov_panel_tf}:
\begin{equation}
    \trfFr{\framePanel}{\frameOdom} = \trfFr{\frameRobot}{\frameOdom} \trfFr{\frameCamera}{\frameRobot} \trfFr{\frameMarker}{\frameCamera} \trfFr{\framePanel}{\frameMarker}
\end{equation}
Consequently, $n$ marker observations lead to $n$ panel pose estimates $\trfFr{\framePanel}{\frameOdom}$ that eventually allow to compute the pose mean which includes mean position and orientation, determined by \emph{spherical linear interpolation (Slerp)} \cite{Shoemake1985}.

\subsection{Panel Handle Pose Estimation ($\mathcal{T}_H$)}
\label{sec:method:panel_comp_det}
Once the panel pose has been estimated, the panel kinematic model can be exploited to approximate the orientations of each component (handles, wheels, valves, etc.).
Accurate orientations are necessary to guarantee reliable manipulation of targets as required by further mission tasks.

\begin{figure}[tb]
	\centering
	\includegraphics[width=0.75\linewidth]{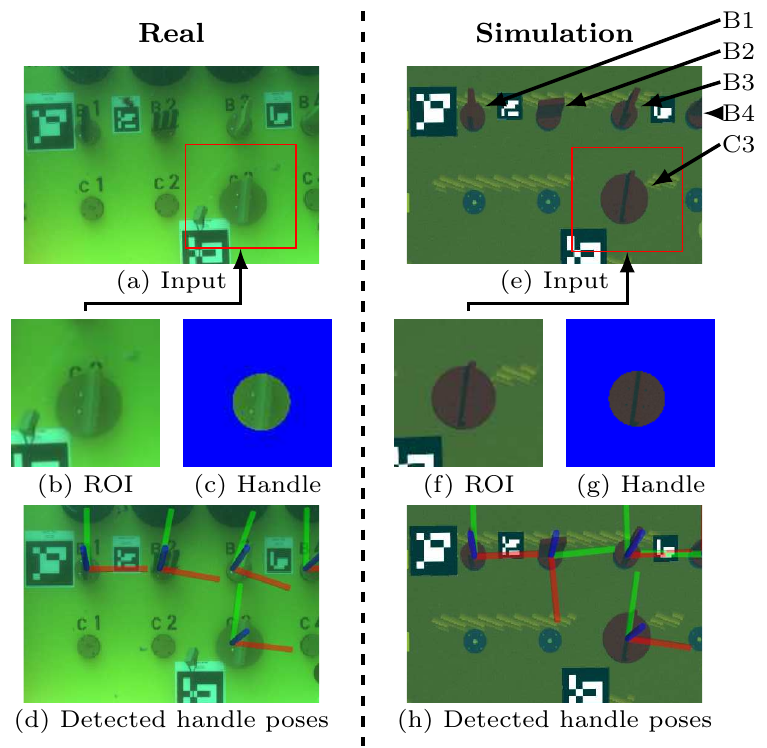}
	\caption{Stages of the handle pose estimation algorithm. It can be tested for identical viewpoints on real ((a)--(d)) and simulation data (e)--(h)).}
	\label{fig:sim_panel_det_state}
\end{figure}

Using the described knowledge base, a \emph{region of interest} (ROI) in form of a 3D oriented bounding box is extracted according to the component model dimensions.
Then an image-based approach estimates the component orientation using the extracted ROI projected into the image space.
Fig.~\ref{fig:sim_panel_det_state}a shows the input image from one monocular view of the real stereo camera, and Fig.~\ref{fig:sim_panel_det_state}b shows the computed ROI of the handle labelled C3.
For precise localization of the handles, the \emph{superellipse-guided active contours segmentation} algorithm \cite{Koehntopp2015} is applied to the image patches representing each handle ROI. This algorithm is a particularly good fit to our use case because the handles are round or ellipsoid-shaped when viewed from a frontal perspective.

Fig.~\ref{fig:sim_panel_det_state}c shows the result of using Fig.~\ref{fig:sim_panel_det_state}b as input for the superellipse-guided active contours segmentation. Subsequently, the handle state given by the lever position can be inferred from the most prominent straight edge in the lever's image. A Canny edge detector is used to detect edge points, followed by a Hough transformation for lines. Fig.~\ref{fig:sim_panel_det_state}d shows the estimated handle poses overlaid over the panel. Based on these estimated orientations, the overall state of the panel is accordingly updated (Fig.~\ref{fig:sim_loop_panel_det_km}).
Then, thanks to the synchronized simulation, the algorithm is tested with an identical camera viewpoint in simulation for comparison, as shown in Fig.~\ref{fig:sim_panel_det_state}e--f; note that, handle pose estimates retrieved from real (Fig.~\ref{fig:sim_panel_det_state}d) and simulated (Fig.~\ref{fig:sim_panel_det_state}h) data may deviate due to different signal-to-noise in the respective data.

To further enhance the robustness of the algorithm when used in the envisioned scenario, a moving average of the detected lever orientations is employed to mitigate the effects of incorrect estimations on single frames. Moreover, both images from the stereo camera are used separately to estimate the handle pose, which gives us two samples from different perspectives at each time instance.

\subsection{Robot Localization ($\mathcal{T}_L$)}\label{sec:localization}
Accurate self-localization of vehicles is a challenging task, especially in the deep-sea domain, due to noisy sensor readings typically based on acoustic devices like Ultra-Short Baseline (USBL) systems, single-beam or multi-beam sonars, Doppler Velocity Log (DVL) or relative readings provided by Inertial Navigation Systems (INS).
Consequently, localization methods rely on multiple modalities to increase reliability~\cite{ChengSurvey2013,Li2016}.
A typical and well-established approach to deal with sensor fusion is the Extended Kalman filter (EKF) \cite{Moore2014} which allows to incorporate these modalities while considering their individual uncertainty.

However, as discussed in the previous section, reliable dexterous manipulation is a requirement in the DexROV scenario.
To ensure robust control of the manipulator arm, accurate robot pose estimates are needed.
Hence, we exploit the panel as a visual landmark again due to its static pose on the seafloor and its visual augmentation with multiple markers.
Once the panel pose has been estimated, the robot pose can be inferred and used as an additional EKF input modality. 
In the following we describe our EKF-based localization system incorporating sensor readings and visual landmarks.

\subsubsection{Sensor Readings}
The robot setup provides a bank of sensors including INS, DVL, and USBL which together allow to gather readings  of the current robot state regarding translation, orientation, linear/angular velocities and accelerations.

\subsubsection{Visual Landmark-Based Localization}
Fig.~\ref{fig:sim_loop_panel_det_real} shows a sample pose estimate of a visual marker, note that the panel is partially observed, used to infer the panel pose though the space transformations shown in Fig.~\ref{fig:rov_panel_tf}.
Now the panel is taken as a fixed landmark and the robot pose \trfFr{\frameRobot}{\frameOdom} can be estimated as follows: 
\begin{equation}
    \trfFr{\frameRobot}{\frameOdom} = \trfFr{\framePanel}{\frameOdom} \trfFr{\frameMarker}{\framePanel} \trfFr{\frameCamera}{\frameMarker} \trfFr{\frameRobot}{\frameCamera}
\end{equation}
where \trfFr{\framePanel}{\frameOdom} is the panel pose in odometry frame, \trfFr{\frameMarker}{\framePanel} is one marker pose in panel frame, \trfFr{\frameCamera}{\frameMarker} is the camera pose w.r.t.\ the marker and \trfFr{\frameRobot}{\frameCamera} is the robot fixed pose w.r.t.\ the camera.
Further on, the means of robot position $\posFrMean{\frameRobot}{\frameOdom}$ and orientation $\quatFrMean{\frameRobot}{\frameOdom}$ w.r.t.\ the odometry frame are estimated from multiple marker detections using \emph{Slerp}.
In addition, a covariance matrix \covFr{\frameRobot}{\frameOdom} for the robot pose is computed: 
\begin{equation}
\covFr{\frameRobot}{\frameOdom}=\mathrm{diag}(\sigma^2_{\pos_{x}},\sigma^2_{\pos_{y}},\sigma^2_{\pos_{z}},\sigma^2_{\quat{q}_{\phi}},\sigma^2_{\quat{q}_{\theta}},\sigma^2_{\quat{q}_{\psi}}).
\end{equation}

The full robot pose estimate $\trfFr{\frameRobot}{\frameOdom} = \langle \posFrMean{\frameRobot}{\frameOdom},\quatFrMean{\frameRobot}{\frameOdom} \rangle$ along with the respective covariance matrix \covFr{\frameRobot}{\frameOdom} is then taken as an input for the localization filter in the final setup. Alternatively, it can be used as a ground truth value to optimize each of its components, i.e. sensor biases and associated covariances.

\subsubsection{Extended Kalman Filter}
In this work, we apply an \emph{Extended Kalman Filter} (EKF) \cite{Moore2014} to estimate the robot pose over time considering a state space consisting of position $x, y, z$, orientation $\phi, \theta, \psi$, translational  $\dot{x}, \dot{y}, \dot{z}$, and angular velocities $\dot{\phi}, \dot{\theta}, \dot{\psi}$ as well as translational accelerations $\ddot{x}, \ddot{y}, \ddot{z}$.
We only incorporate direct sensor measurements to the EKF, no integrated or differentiated values. INSs produce angular and linear accelerations, a DVL provides position outputs in form of altitude readings and linear velocities, and the mentioned landmarks are incorporated as pose readings. To increase the localization filter robustness, obvious outliers from sensor readings are rejected heuristically, and the pose inputs inferred from visual markers are tuned based on our experimental results.

\section{Experimental Evaluation} \label{sec:experiment}
In the following experiments we exploit the SIL concept with real-world data recorded during field trials in Marseille, France, in July 2017. The goal is to demonstrate the effectiveness of the proposed method to benchmark critical system components and several mission tasks $\mathcal{T}_i$ in deep-sea operations.
This data contains sequences where the robot was used to verify the integrity of the testing panel and expected handle positions at up to 100 meters below sea level. 

In the first experiment we perform an \emph{environment condition adaptation} task ($\mathcal{T}_E$) to find the best possible simulation setup with respect to environment conditions. Next, we evaluate the performance of three benchmarking tasks: \emph{panel pose estimation} ($\mathcal{T}_P$), \emph{panel handle pose estimation} ($\mathcal{T}_H$) and \emph{robot localization} ($\mathcal{T}_L$).

\subsection{Environment Condition Adaptation ($\mathcal{T}_{E}$)}
\label{Environment Condition Adaptation}

\begin{figure}
	\centering
	\captionsetup{justification=centering}
	\subcaptionbox{\\$m(\mathcal{T}_E)=1.0$\label{fig:image_adaptation_real}}{
		\includegraphics[height=0.28\columnwidth]{real_marker_scaled.png} 
	}
	\subcaptionbox{$m(\mathcal{T}_E)=0.32$\label{fig:image_adaptation_sim1}}{
		\includegraphics[height=0.28\columnwidth]{sim_no_noise.png} 
	}
	\subcaptionbox{$m(\mathcal{T}_E)=0.48$\label{fig:image_adaptation_sim2}}{
		\includegraphics[height=0.28\columnwidth]{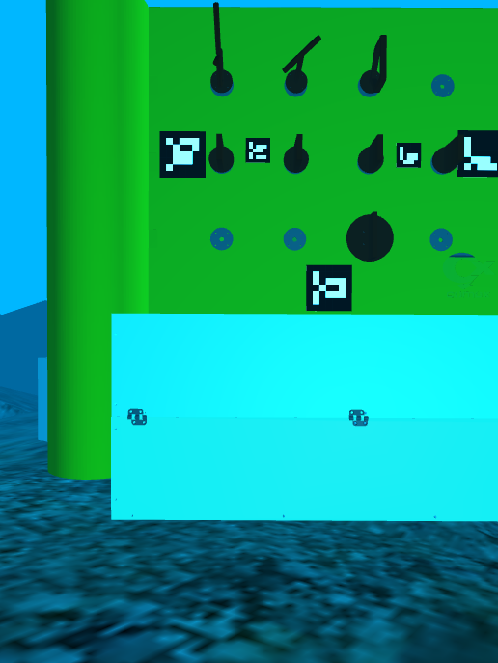} 
	}
	\subcaptionbox{$m(\mathcal{T}_E)=0.65$\label{fig:image_adaptation_sim3}}{
		\includegraphics[height=0.28\columnwidth]{sim_noisy.png} 
	}
	\captionsetup{justification=justified}
	\caption{Simulated image adaptation to real environment conditions. Images correspond to
	(\subref{fig:image_adaptation_real}) real-world, (\subref{fig:image_adaptation_sim1}) default and (\subref{fig:image_adaptation_sim2}--\subref{fig:image_adaptation_sim3}) light adapted simulation.}
	\label{fig:uw_noisy_images}
\end{figure}

In this task the simulated environment conditions $\mathcal{E}$ are tuned to reflect real environment conditions with high fidelity. We focus on adapting the light behavior to replicate the underwater camera distortions, i.e. light and color attenuation. The simulated stereo camera applies an exponential attenuation on the pixel intensity as described in \cite{Manhaes2016}:
\begin{equation} 
i_{c}^{*}=i_{c}e^{-za_{\mathrm{c}}}+(1-e^{-za_{\mathrm{c}}})b_{c}\quad\forall c\in\{R, G, B\}
\label{eq: uw_camera} 
\end{equation}
where $i_c$ and $b_c$ correspond to the pixel and background intensity value for color channel $c$, and  $a_c$ is a color-dependent attenuation factor. The attenuation depends on the distance $z$ to the object projected on the camera pixel, which is extracted directly from the simulator depth-camera plugin.

We define the measure $m(\mathcal{T}_E)$ equivalent to the Feature Simularity Index (FSIM) \cite{Zhang2011_FSIM} image quality measure between synchronized real and simulated images. 50 images recorded from the field trial were used to heuristically determine the most adequate light parameters for Equation~\ref{eq: uw_camera}; these images include different distances to the panel and perspectives. Fig.~\ref{fig:uw_noisy_images} shows some adapted image instances and their respective $m(\mathcal{T}_E) \in [0,1]$. Fig.~\ref{fig:image_adaptation_sim3} shows the optimized adapted image under environment conditions $\mathcal{E}^*$, which are used in the next experiments to find the \emph{expected system performance} in simulation.

\subsection{Panel Pose Estimation ($\mathcal{T}_P$)}
\label{exp:panel_detection}

\begin{figure}
	\centering
	\resizebox{0.9\linewidth}{!}{\input{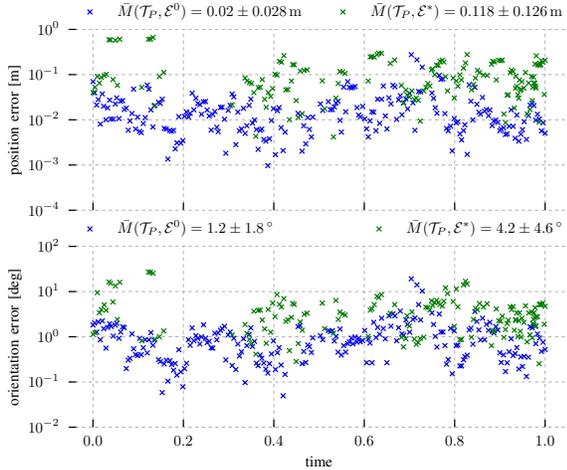}}
	\caption{Panel detection error $M(\mathcal{T}_P)$ for noise-free \textcolor{blue}{$\mathcal{E}^0$} and underwater \textcolor{DarkGreen}{$\mathcal{E}^*$} environment conditions}
	\label{fig:panel_detection_error}	
\end{figure}

This first benchmarking task $\mathcal{T}_P$ evaluates the accuracy of the panel pose estimation, as it is the starting point for other tasks like handle pose estimation. This consequently validates the robustness of the used visual markers.

In simulation the robot navigates as in Fig.~\ref{fig:real_gt_vs_marker_vs_ekf_trajectory}; the trajectory was computed by extracting the robot poses given by the detected markers on real data and using them as waypoints in simulation.
In this way, the same visual perspectives as in the field trial are obtained which represent a common routine trajectory given by the robot operators. Thus, we can determine the expected error as the difference between the ground-truth panel pose in simulation $\panelPoseSimulation$ and the panel pose determined from marker detection $\panelPoseMarker$:
\begin{align} 
m(\mathcal{T}_{P},\mathcal{E})= & \panelPoseError{S}{M} \nonumber \\
= & \langle \panelPositionError{S}{M} , \panelOrientationError{S}{M} \rangle
\label{eq:pose_error_measure} 
\end{align}
where $\panelPositionError{S}{M}$ is the Euclidean distance between positions and $\panelOrientationError{S}{M}$ is the minimal geodesic distance between orientations \cite{Huynh2009} under conditions $\mathcal{E}$. 

Fig.~\ref{fig:panel_detection_error} shows the mean $\bar{M}(\mathcal{T}_P,\mathcal{E})$ and standard deviation $\sigma(M(\mathcal{T}_{P},\mathcal{E}))$ for all panel observations under noise free \textcolor{blue}{$\mathcal{E}^0$} and underwater-like conditions \textcolor{DarkGreen}{$\mathcal{E}^*$}. 
As expected, our method has high accuracy under noise-free environment, and underwater image distortions decrease the accuracy and number of detections. 
With \textcolor{DarkGreen}{$\mathcal{E}^*$} conditions, we can expect a translation and orientation error of \SI{0.118}{m} and \SI{4.2}{^{\circ}} respectively. 
For visual survey and navigation tasks this error is minimal. It can be overcome by image registration methods and the variance $\sigma^2(M(\mathcal{T}_{P},\mathcal{E}^*))$ can be used to fine-tune the robot pose covariance matrix \covFr{\frameRobot}{\frameOdom} (Section~\ref{sec:localization}) to improve localization, as it is an estimation of pose precision. 
For manipulation tasks, applying a moving average over the outputs produces good results as explained in Section~\ref{sec:method:panel_comp_det}. Tasks $\mathcal{T}_H$ and $\mathcal{T}_L$  show this in the next experiments.             

\subsection{Panel Handle Pose Estimation ($\mathcal{T}_H$)}
\label{exp: handle state detection}
\begin{figure}
	\centering
	\includegraphics[width=0.9\linewidth]{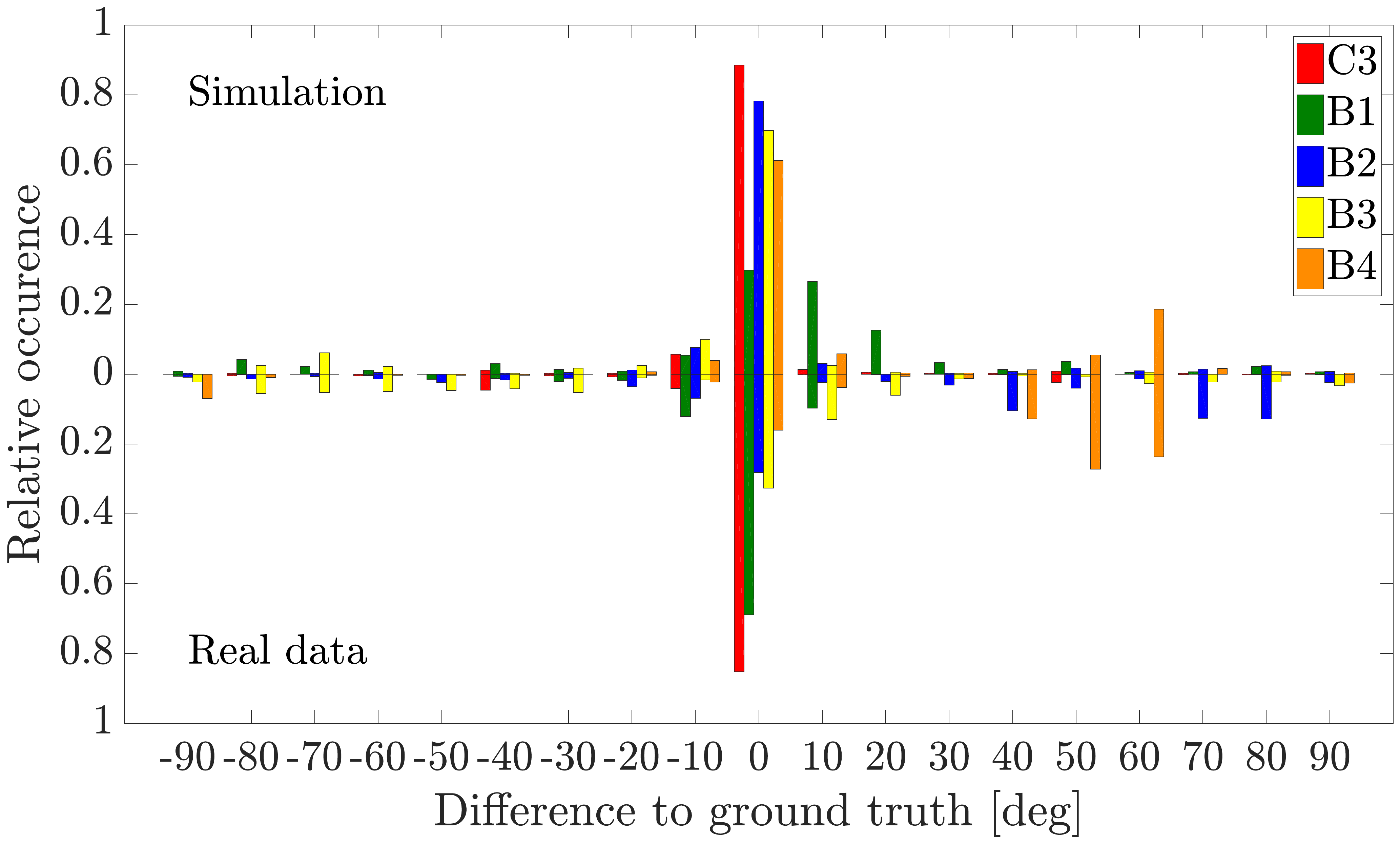}
	\caption{Normalized histograms of handles orientation error $M(\mathcal{T}_{H})$ for simulated and real data}
	\label{fig:handleStateDetectionResult}
\end{figure}
The previously described handle pose estimation was tested with the SIL framework, synchronizing the real and simulated camera point of view. Fig.~\ref{fig:handleStateDetectionResult} shows the normalized histograms of the error $M(\mathcal{T}_{H})$ between the estimated handle orientation and ground truth from simulated and real data; note that, ground truth handle orientations are retrieved from visual inspection. 
In the course of the development, B1 and B2 have been modified to enhance the estimation accuracy by coloring the respective lever black in order to investigate contrast benefits for edge-based image algorithms as the one used (see Figs.~\ref{fig:sim_panel_det_state}a and \ref{fig:sim_panel_det_state}e). 
This modification causes a discrepancy between simulated and real data, therefore B1 and B2 are disregarded in the results shown in Fig.~\ref{fig:handleStateDetectionResult}.
In general, the results in Fig.~\ref{fig:handleStateDetectionResult} show that the handle orientation estimates are predominantly within a low -10$^\circ$ to 10$^\circ$ error range.
Depending on viewpoint, handle type and orientation, outliers can be observed for B4 in the range of 60$^\circ$ to 70$^\circ$.

\subsection{Robot Localization ($\mathcal{T}_L$)}
\label{exp:localization}
The final task $\mathcal{T}_L$ to benchmark with our SIL methodology is the localization method described in Section~\ref{sec:localization}. First we validate the use of visual landmarks in the localization filter through simulation under the found $\mathcal{E}^*$ conditions, then with real data we compare the task performance with and without the use of visual landmarks, and finally we tune the EKF parameters based on the results from tasks $\mathcal{T}_{L1}$ and $\mathcal{T}_{P}$.

\subsubsection{$\mathcal{T}_{L1}$ -- localization in simulation}
\begin{figure*}
    \centering
    \begin{subfigure}[b]{.48\textwidth}
        \resizebox{.8\linewidth}{!}{\input{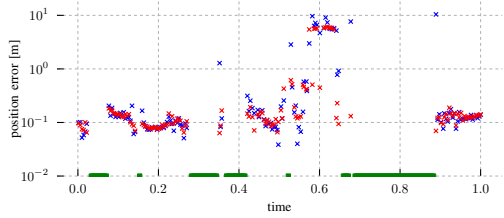}}
        \caption{Position errors \textcolor{red}{$\robotPositionError{S}{M}$} and \textcolor{blue}{$\robotPositionError{S}{F}$}.\\No marker detected for sampling times marked \textcolor{DarkGreen}{green} \label{fig:real_gt_vs_marker_vs_ekf_position}}
        \resizebox{.8\linewidth}{!}{\input{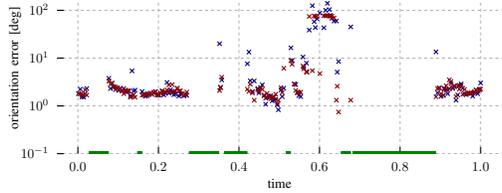}}
        \caption{Orientation errors \textcolor{DarkRed}{$\robotOrientationError{S}{M}$} and \textcolor{DarkBlue}{$\robotOrientationError{S}{F}$}.\\No marker detected for sampling times marked \textcolor{DarkGreen}{green} \label{fig:real_gt_vs_marker_vs_ekf_orientation}}
    \end{subfigure}
    \begin{subfigure}[b]{.48\textwidth}
        \resizebox{.9\linewidth}{!}{\input{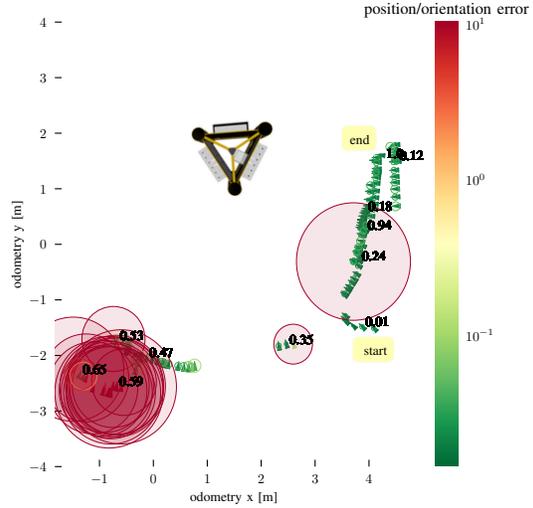}}
        \caption{Robot poses (triangles) with orientation error $\robotOrientationError{S}{F}$ (triangle color) and position error $\robotPositionError{S}{F}$ (circle color, log-scaled circle radius). Some time marks are shown next to some poses for reference\label{fig:real_gt_vs_marker_vs_ekf_trajectory}}
    \end{subfigure}
	\caption{$\mathcal{T}_{L1}$ results: position and orientation errors between ground-truth robot pose and marker-based / localization filter-based robot pose estimates, while the robot moves around the panel on a trajectory recorded in field trials}
  \label{fig:results_localization_real}
\end{figure*}

In the case of real-world underwater localization, no accurate ground-truth sensing is available. For this reason, the performance of the proposed localization filter that integrates visual landmarks into the EKF has to be tested in simulation first; and deployed afterwards in the field. 

In this task $\mathcal{T}_{L1}$ the simulated robot again follows the trajectory shown in Fig.~\ref{fig:real_gt_vs_marker_vs_ekf_trajectory}. During this movement, the ground-truth robot pose in simulation $\robotPoseSimulation$ is recorded alongside the robot pose determined through the detected marker pose $\robotPoseMarker$ and the localization filter $\robotPoseFilter$. Note that the EKF receives only the visual landmark-based pose estimates to prove that it converges to ground truth.

The pose estimate error of $\robotPoseMarker$ and $\robotPoseFilter$ with respect to simulation ground truth are shown in Fig.~\ref{fig:results_localization_real}, denoted as $m_{S,M}(\mathcal{T}_{L1})=\robotPoseError{S}{M}$ and $m_{S,F}(\mathcal{T}_{L1})=\robotPoseError{S}{F}$ as in Equation~\ref{eq:pose_error_measure}.
The trajectory in Fig.~\ref{fig:real_gt_vs_marker_vs_ekf_trajectory} and the detailed error breakdown in Fig.~\ref{fig:real_gt_vs_marker_vs_ekf_position}--\subref{fig:real_gt_vs_marker_vs_ekf_orientation} show that whenever no marker has been detected for a while, the EKF error increases significantly on the next reading, but then quickly re-converges towards ground truth. On parts of the trajectory where markers are visible constantly, the localization error decreases satisfactorily below \SI{0.3}{m}/\SI{3}{^{\circ}} e.g. between time marks $0.1$ and $0.25$.

\subsubsection{$\mathcal{T}_{L2}$ -- real-world localization using only navigation sensor data}
In order to get a baseline to compare the performance of the localization filter when integrating visual landmarks, in this subtask only DVL and INS measurements are used as inputs. This is also done because our focus is on tuning the use of visual markers in the EKF because navigation sensors are not integrated in the simulation at this development stage.

To evaluate this and the next tasks, see Table~\ref{table:localization_tasks}-\ref{table:localization_measures}, we use the robot pose estimate given by the marker $\robotPoseMarker$ as reference ground truth, and compute the measure $m_{M,F}(\mathcal{T}_{Li})=\robotPoseError{M}{F}$ plus the \emph{lag-one autocorrelation} $m_{A}(\mathcal{T}_{Li})=\sum_{t}\robotPoseFilter (t)\robotPoseFilter (t-1)$ on the EKF-predicted poses. $m_{A}(\mathcal{T}_{Li})$ is a measure of trajectory smoothness, important to prevent the robot from performing sudden jumps that can interfere with manipulation tasks. 

\subsubsection{$\mathcal{T}_{L3},\mathcal{T}_{L4},\mathcal{T}_{L5}$ -- real-world localization with visual markers}
\begin{figure}
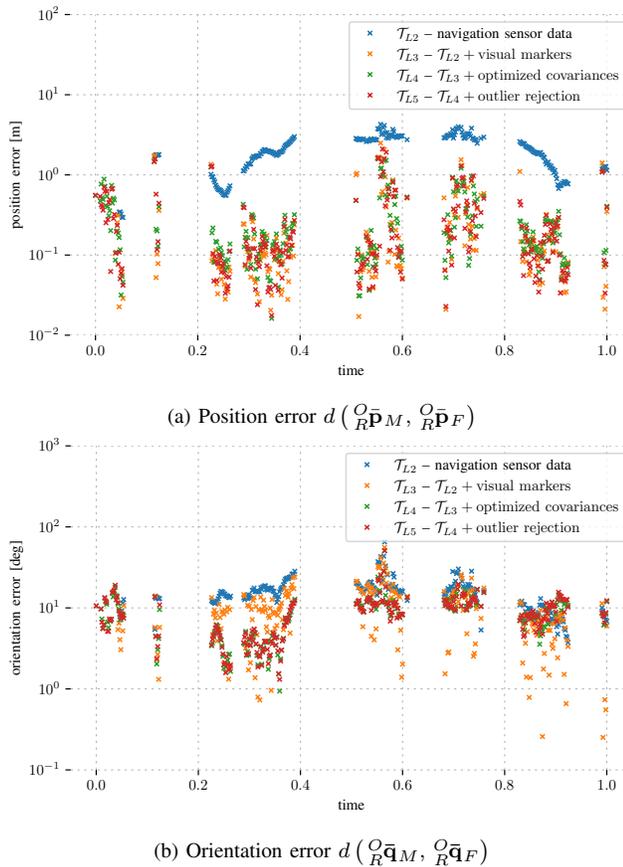

	\centering
	\captionsetup{justification=centering}
	\begin{subfigure}[b]{\linewidth}
		\centering
		\resizebox{\linewidth}{!}{\input{localization_error_pos_log.pgf}}
		\caption{Position error \robotPositionError{M}{F}\label{fig:real_localization_pos_results}}
	\end{subfigure}
	\begin{subfigure}[b]{\linewidth}
		\centering
		\resizebox{\linewidth}{!}{\input{localization_error_or_log.pgf}}
		\caption{Orientation error \robotOrientationError{M}{F}\label{fig:real_localization_or_results}}
	\end{subfigure}
	\caption{Localization benchmark tasks $\mathcal{T}_{Li}$ results}
	\label{fig:localization_plot_results}
\end{figure}

In these tasks, we show the localization results using all sensor data recorded in field trials along with visual landmark-based pose estimates.
A description of the respective tasks is given in Table \ref{table:localization_tasks}.
The corresponding results are shown in Table~\ref{table:localization_measures} and Fig.~\ref{fig:localization_plot_results}.
\begin{table}
	\centering
	\scriptsize
	\captionsetup{justification=centering}
	\caption{Description of localization tasks $\mathcal{T}_{Li}$}\label{table:localization_tasks}
	\begin{adjustbox}{max width=.95\linewidth}
	\begin{tabularx}{\linewidth}{lX}
		\toprule
		\textbf{Task} & \textbf{Description} \\ \midrule
		$\mathcal{T}_{L2}$ & EKF with real-world data and only navigation sensors \\
		$\mathcal{T}_{L3}$ & EKF with real-world data, using navigation sensors and  visual markers (default parameters) \\
		$\mathcal{T}_{L4}$ & $\mathcal{T}_{L3}$, plus covariance \covFr{\frameRobot}{\frameOdom} of the robot pose estimates from marker detections adjustment with results from task $\mathcal{T}_{P}$, i.e.\ using $(\SI{0.126}{m})^2$ and $(\SI{4.6}{^{\circ}})^2$ (see Fig.~\ref{fig:panel_detection_error}) as diagonal values for single marker detections \\
		$\mathcal{T}_{L5}$	& $\mathcal{T}_{L4}$, plus rejection of pose estimates whose distance $\robotPoseError{M}{F}$ to the current prediction are greater than \SI{1}{m} and \SI{12}{^{\circ}}; according to $\mathcal{T}_{L1}$ and  Fig.~\ref{fig:real_gt_vs_marker_vs_ekf_position}--\ref{fig:real_gt_vs_marker_vs_ekf_orientation} \\ \bottomrule
	\end{tabularx}
	\end{adjustbox}
\end{table}

\begin{table}
	\centering
	\captionsetup{justification=centering}
	\caption{Tasks $\mathcal{T}_{Li}$ measure results}\label{table:localization_measures}
	\resizebox{\linewidth}{!}{%
		\begin{tabular}{lS[table-format=2.2(2)]S[table-format=1.2(2)]S[table-format=1.2(2)]S[table-format=1.2(2)]}
			\toprule
			& $\mathcal{T}_{L2}$ & $\mathcal{T}_{L3}$ & $\mathcal{T}_{L4}$ & $\mathcal{T}_{L5}$ \\ \midrule
			$\bar{m}_{M,F}(\mathcal{T}_{Li}\langle \bar{\mathbf{p}} \rangle )\mathrm{[m]}$ & 2.11 \pm 0.94 & 0.26 \pm 0.39 & 0.29 \pm 0.32 & 0.28 \pm 0.35\\
			$\bar{m}_{M,F}(\mathcal{T}_{Li}\langle \bar{\mathbf{q}} \rangle )\mathrm{[deg]}$ & 15.59 \pm 7.33 & 10.24 \pm 7.57 & 8.82 \pm 5.17 & 8.86 \pm 5.19\\
			$m_{A}(\mathcal{T}_{Li})$ & 0.95 & 0.72 & 0.91 & 0.94 \\ \bottomrule
		\end{tabular}
	}
\end{table}

As expected, the EKF instance with only navigation sensors as an input ($\mathcal{T}_{L2}$ -- \textcolor[rgb]{0,0.47,0.73}{blue} in Fig.~\ref{fig:localization_plot_results}) bears the largest error. Integrating the visual markers ($\mathcal{T}_{L3}$ -- \textcolor[rgb]{1,0.56,0}{orange}) significantly reduces the error and increases the number of jerky movements while navigating; this results in the worst $m_{A}(\mathcal{T}_{Li})$. 
Finally, we show that, based on previous tasks $\mathcal{T}_{P}$ and $\mathcal{T}_{L1}$ developed through our SIL methodology, the localization performance can be optimized adjusting the pose estimates covariances ($\mathcal{T}_{L4}$ -- \textcolor[rgb]{0,0.6,0.19}{green}) and rejecting outliers ($\mathcal{T}_{L5}$ -- \textcolor[rgb]{0.7,0.03,0}{red}). $\mathcal{T}_{L4}$ and $\mathcal{T}_{L5}$ yield similar accuracies, but the latter achieves the smoothest navigation trajectory. 
Certainly, as more sensors are integrated in simulation, performance can be further enhanced through simulation in the loop.   

\section{Conclusion}
\label{sec:conclusion}
Deep-sea robotic operations are cost intensive, and demand robustness and high reliability under harsh conditions.
Measures have to be taken to guarantee the safety of the crew and the equipment.
This includes not only robust pose estimation and localization algorithms, but also a versatile development framework including a realistic testbed.

In this work we presented a simulation in the loop (SIL) procedure that incorporates real observations into the simulation in a seamless manner by synchronization of simulated conditions with real-world data. 
Consequently, the components development progress can be instantaneously verified and benchmarked under simulated and real conditions. 
In our experimental evaluation we showed the benefit of the presented SIL approach on the DexROV research project.
We were able to analyze and optimize critical components like robot localization considering the components' behavior under various environmental and spatial conditions.

\bibliographystyle{IEEEtran}
\bibliography{bibliography}

\end{document}